\documentclass[10pt,twocolumn,letterpaper]{article}

\usepackage{wacv}              

\usepackage{tikz}
\usepackage{graphicx}
\usepackage{booktabs}
\usepackage{amsfonts}
\usepackage{todonotes}
\usepackage{amsmath}

\newcommand{\figref}[1]{Fig.\,\ref{#1}}

\newcommand{\tabref}[1]{Tab. \,\ref{#1}}

\DeclareMathOperator*{\argmin}{argmin}

\def\D#1{\textcolor{black}{#1}}

\usepackage{pifont}
\usepackage{multirow}

\definecolor{wacvblue}{rgb}{0.21,0.49,0.74}
\usepackage[pagebackref,breaklinks,colorlinks,allcolors=wacvblue]{hyperref}

\def\wacvPaperID{696} 
\def\confName{WACV}
\def\confYear{2027}

\title{Structure-aware Riemannian Growth Fields for 4D Plant Modeling}

\author{Meng-Yu Jennifer Kuo\\
Nara Women's University, Japan\\
\and
Ryo Kawahara\\
Kyoto University, Japan\\
}

\begin{document}
\twocolumn[{%
\maketitle
\centering
\vspace{-20pt}
\begin{center}
      \centering
    \includegraphics[clip, trim=0.0cm 0.0cm 0.0cm 0.0cm, width=1\linewidth]
    {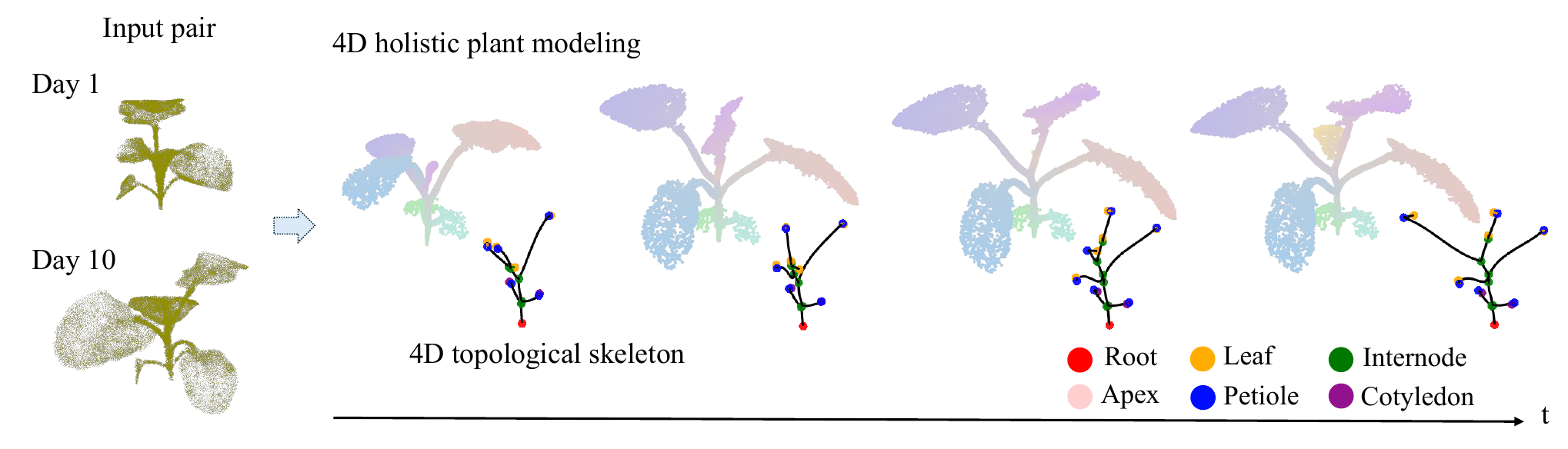}
    \vspace{-19pt}
    \captionof{figure}{Our method enables spatio-temporal, dense and holistic 4D plant generation from as few as two sparse timepoint observations. The results show that the method can track correspondences as time progress, which would otherwise be hard to achieve with conventional methods. 
    Nodes on the topological skeleton are color-coded by their tracking identities to demonstrate structural consistency.
    }
\label{fig:teaser}
\end{center}	
}]

\begin{abstract}
\D{In this paper, we introduce a novel framework for 4D plant growth modeling that reconstructs the continuous geometric and topological evolution of plants from sparse temporal observations. Existing methods mainly rely on dense registration, yet reliable dense sequences are hard to obtain due to scanning constraints and self-occlusions, leaving these approaches struggling under large temporal gaps where rapid organ emergence violates local rigidity. To overcome this, we bridge these gaps by formulating plant morphogenesis as a continuous procedural process on a structure-aware Riemannian growth field; this jointly models topology evolution and geometric deformation, preserving botanical hierarchies and stable spatio-temporal correspondences across distant timepoints. Our key idea is to ground symbolic growth rules within a continuous geodesic flow, where organ development follows biologically modulated trajectories that preserve structural coherence under topological changes. We further contribute a 10-day dual-species dataset with dense geometric and semantic annotations. Experiments demonstrate that our method accurately tracks individual organ growth over time and significantly outperforms state-of-the-art baselines in both geometric accuracy and correspondence consistency.}
\end{abstract}
    

\section{Introduction}
\label{sec:intro}

Plant modeling plays an increasingly important role in applications ranging from precision agriculture and autonomous greenhouse management to environmental biophysical simulation. Recovering high-fidelity plant geometry remains challenging due to intricate branching structures, severe self-occlusions, and texture sparsity. Traditional photogrammetry and LiDAR methods often struggle to capture these thin, overlapping organs, whereas recent 3D Gaussian Splatting (3DGS) approaches \cite{ojo2024splanting, zhang2025wheat3dgs, yang2025gaussianplant, hartley2025plantdreamer} have demonstrated promising results through explicit surface primitives and high-quality rendering. Nevertheless, existing frameworks largely treat plants as static entities. Modeling continuous plant growth is essential for understanding developmental dynamics, yet it remains fundamentally difficult due to large non-rigid deformations and persistent topological changes inherent in biological morphogenesis.

\D{Recent work in 4D plant modeling seeks to reconstruct continuous growth from discrete data, yet acquiring dense temporal captures is often infeasible due to hardware constraints and severe self-occlusions over cultivation cycles. Under such sparse constraints, registration methods \cite{pan2021multi, adebola2025growsplat, chebrolu2020spatio} rely on local geometric approximations, failing across distant timepoints where abrupt organ emergence violates the local-rigidity assumption. While skeletal frameworks \cite{khanam2024riemannian} introduce formal shape metrics, their optimization lacks biological constraints and is prone to local minima that violate temporal causality without dense inputs, causing branch sliding and corrupted correspondences. A separate line of work injects botanical priors through rule-based grammars 
\cite{prusinkiewicz1999system, lintermann1999interactive}: classical differential L-systems \cite{prusinkiewicz1993animation} \D{model growth via forward simulation of hand-specified symbolic rules in flat Euclidean space}, whereas recent data-driven approaches \cite{lee2023latent, ghrer2026learning} predict static geometry from 3D data---neither optimizing a developmental trajectory over the intrinsically curved manifolds of shape space.}

\D{This raises a fundamental question: \textit{Can we reconstruct continuous, biologically plausible development given reliable data at only a few sparse timepoints?} In this paper, we introduce a novel framework for reconstructing continuous 4D plant morphogenesis from as few as two sparse temporal observations (\figref{fig:teaser}). Rather than treating growth as registered snapshots or static grammar predictions, we jointly model topology evolution and geometric deformation, enabling stable spatio-temporal correspondences throughout development.
Our key idea is to formulate plant development as a continuous procedural process on a structure-aware Riemannian growth field. By replacing discrete Euclidean rules with continuous geodesic dynamics, organ elongation and expansion follow energetically optimal trajectories. 
This bridges high-level botanical hierarchy and geometric evolution, yielding a unified 4D representation that preserves structural identity even as topology changes.}

More specifically, we represent the evolving plant skeleton as a dynamic graph in which the birth of each node is triggered by a symbol-specific wait interval relative to its parent. To ensure realistic deformation, we derive a continuous growth velocity field that minimizes a morphological smoothness energy across a stratified manifold, where new structural degrees of freedom emerge at birth events. The velocity field is further governed by a sigmoid growth modulation that enforces biologically plausible maturation limits. Building on these structural trajectories, we generate a holistic 4D representation of stems and leaves, whose surface evolution follows anisotropic expansion aligned with intrinsic growth directions.

We collect a 10-day dual-species dataset comprising daily growth measurements of bean and broccoli specimens. This dataset provides comprehensive 4D annotations spanning dense geometry, extracted skeletal topologies, and semantic labels for individual organs.
Experimental results demonstrate that our method reconstructs continuous growth with stable spatio-temporal correspondences, improved geometric fidelity, and topological consistency\D{---outperforming conventional methods that struggle under sparse, large-displacement settings.}




\section{Related Works}
\label{sec:related}

\subsection{Static Plant 3D Reconstruction}
Recent research on plant reconstruction has transitioned from traditional approaches such as Structure-from-Motion (SfM) and LiDAR \cite{zhang2024cucumber, wang20223dphenomvs, tang2024lidar, gene2020fruit} toward high-fidelity neural representations \cite{adebola2025growsplat, yang2025gaussianplant, zhang2025wheat3dgs, liu2025boxplant, ojo2024splanting, hartley2025plantdreamer, yang2025neuraleaf, isokane2018probabilistic} that allow for explicit surface extraction.
In particular, 3D Gaussian Splatting (3DGS) \cite{kerbl3Dgaussians} has emerged as a pivotal technique for representing botanical scenes through anisotropic Gaussian primitives. \textit{Splanting} \cite{ojo2024splanting} introduced a fast-capture phenotyping pipeline based on 3DGS, while Zhang \etal \cite{zhang2025wheat3dgs} integrated 3DGS with the Segment Anything Model (SAM) \cite{kirillov2023segment} for automated wheat head segmentation and trait extraction. 
Despite their exceptional geometric detail, these representations treat the plant as a static snapshot and lack a temporal mechanism to model the continuous growth processes of the specimen.


\subsection{Procedural Plant Growth Modeling}
\D{Most botanical frameworks simulate organs \cite{kim2017procedural, lu2009venation} or global architecture 
\cite{prusinkiewicz1999system, lintermann1999interactive, mvech1996visual, prusinkiewicz2012algorithmic} 
via manual biological priors. To model continuous temporal changes, classical differential L-systems \cite{prusinkiewicz1993animation} combine symbolic rules with differential equations; however, they operate as hand-specified forward simulations in flat Euclidean space, failing to capture how biological forms develop on the intrinsically curved manifolds of shape space \cite{godin2025riemannian}. While recent data-driven methods \cite{lee2023latent, ghrer2026learning} learn grammars or parameterized representations directly from 3D target observations, they rely on a discrete sequence-generation paradigm, producing static geometry without optimizing the deformation trajectory. In contrast, we ground data-driven topological rule generation within a continuous geodesic flow on a stratified Riemannian manifold, optimizing across sparse temporal endpoints into an energy-optimal, biologically consistent growth trajectory.}

\begin{figure*}[t]
\begin{center}
\includegraphics[clip, trim=0cm 0cm 0cm 0cm, width=0.88\linewidth]{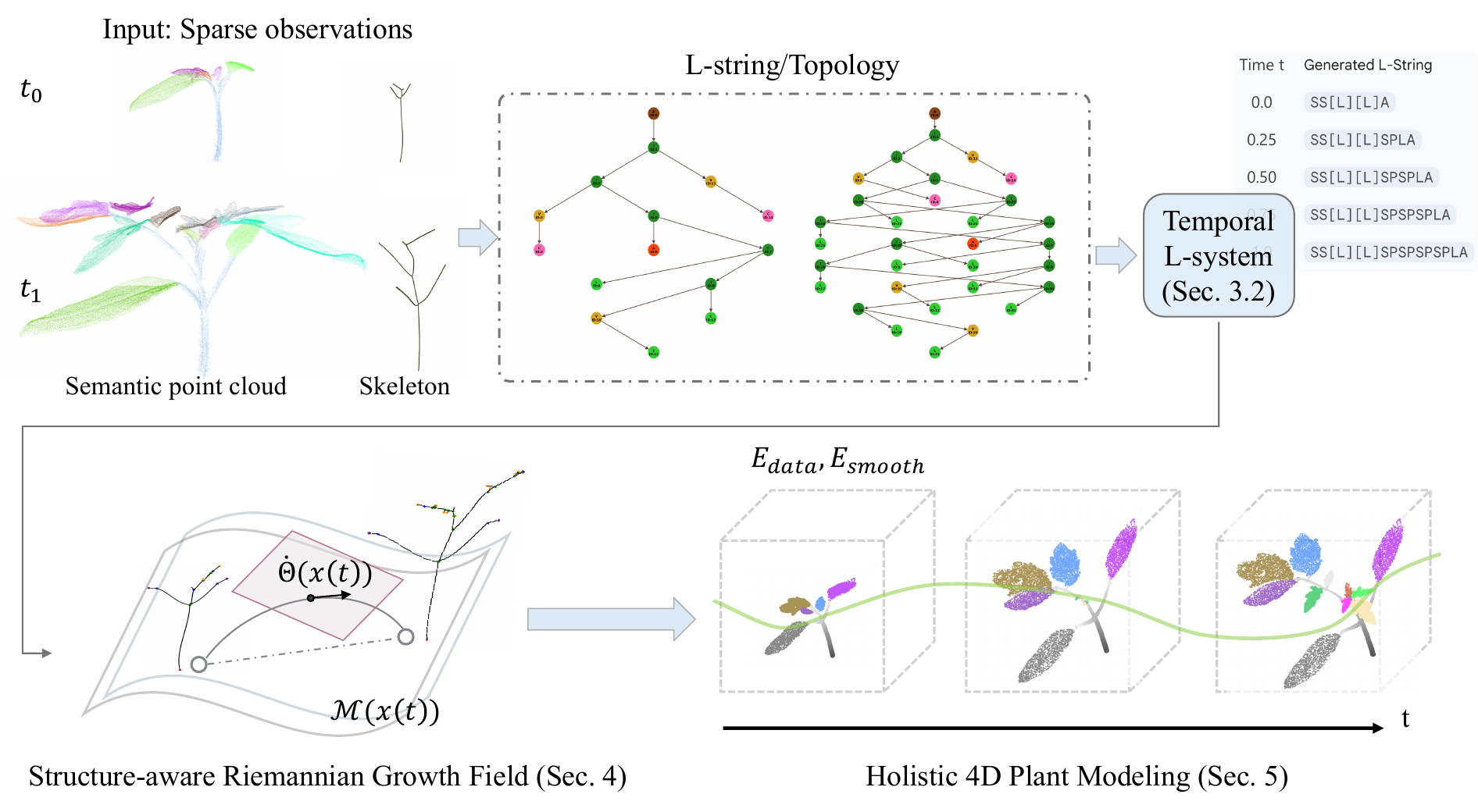}
\vspace{-15pt}
\end{center}
   \caption{Overview of our 4D Plant Modeling Pipeline. Given as few as two sparse temporal semantic point clouds, we first recover the structural scaffold and perform continuous topological growth inference to optimize a temporal L-system $L(t)$ (Sec. 3). These symbolic transitions are then grounded in a structure-aware Riemannian growth field on a stratified manifold $\mathcal{M}(x(t))$ (Sec. 4). By integrating a continuous growth velocity field $\dot{\Theta}(x(t))$ governed by morphological smoothness and sigmoid maturation priors, our framework generates a holistic 4D representation with dense spatio-temporal correspondences (Sec. 5).}
\label{fig:pipeline}
\end{figure*}


\subsection{4D Plant Modeling and Tracking}
\D{Reconstructing temporally consistent plant models is challenging due to the non-rigid deformations and topological changes inherent in biological growth. 
Geometry-based registration frameworks \cite{pan2021multi, adebola2025growsplat, chebrolu2020spatio} rely on local approximations and require dense temporal sampling, failing across wide observation gaps where rapid organ emergence violates local rigidity. 
The closest work to ours is the Riemannian framework by Khanam \etal \cite{khanam2024riemannian}, which uses Square Root Velocity Function Trees (SRVFT) for non-rigid alignment. Though providing a rigorous metric for skeletal trajectories, its shape-space optimization lacks biological priors. Under sparse inputs, this geometric matching struggles with the search space of discrete organ appearance and is prone to local minima that violate temporal causality, causing branch sliding and corrupted correspondences. In contrast, our framework couples botanical hierarchies with continuous Riemannian flow on a stratified manifold, leveraging symbolic growth rules to recover intermediate birth events and maintain dense, biologically faithful spatio-temporal correspondences from sparse observations.}

\section{Spatio-Temporal Structural Inference}

To model morphogenesis as a continuous trajectory, we first establish a consistent topological transition between discrete observations. This section details the reconstruction of a structural scaffold from raw sensor data and the subsequent inference of the temporal rules governing its evolution. By mapping unstructured point clouds onto a dynamic symbolic representation, we provide the discrete boundary conditions required for the high-dimensional Riemannian growth fields described in Section 4. In Fig.~\ref{fig:pipeline}, we illustrates the overall pipeline of our 4D plant modeling.

\subsection{Topological Recovery and Graph Initialization}
To establish a rigorous basis for 4D growth inference, given two semantic point clouds $P_{t_0}$ and $P_{t_1}$, we first recover the underlying structural skeletal representation. We employ Laplacian-based contraction (LBC) \cite{meyer2023cherrypicker} to derive an initial set of 1D curves that approximate the plant's medial axis.
To mitigate structural discontinuities arising from sensor occlusions, we perform a directional search along the local tangent of orphaned terminal nodes. This process yields a canonical skeletal graph $G = (V, E)$ where each vertex is assigned a semantic attribute (e.g., internode, leaf).

By serializing $G$ through a recursive depth-first traversal, we generate the bracketed L-string that defines the plant's symbolic state. This initialization transforms the unstructured spatial data into a topological scaffold, providing the discrete boundary conditions necessary for the subsequent temporal inference.

\subsection{Continuous Topological Growth Inference}


Given L-strings at the observed time instances, we infer the production rules governing the transition between these states. \D{Whereas prior L-systems \cite{prusinkiewicz121996systems, prusinkiewicz1993animation} rely on hand-specified symbolic rules, ours are recovered directly from real data, modeling topological evolution as continuous, time-dependent transitions. At this stage, we treat the L-system as a symbolic scaffold devoid of geometric attributes such as branch length.}

\subsubsection{Temporal L-system}
We represent the time-varying symbolic string of the evolving plant as a function of time $L(t)$. 
We define the maturation birth time $\tau_v$ for each node $v$ through a recursive temporal propagation from its parent $v_p$:
\begin{equation}
\tau_v = \tau_{v_p} + \Delta\tau_{\mu(v)} \,,
\end{equation}
where $\Delta \tau_{\mu(v)}$ is a \textit{symbol-specific wait interval} modeling organ-specific dormancy. 
In this formulation, the emergence of each node is no longer constrained to global discrete generation steps. Instead, each node's birth is triggered by the completion of its parent's growth, forming a sequential chain of topological events.

\subsubsection{Dynamic Graph Representation}
The birth time $\tau_v$ defines the evolution of the semantic plant graph $G(t) = (V(t), E(t))$. At any time $t$, the graph consists of elements from the potential structure $(\mathcal{V}, \mathcal{E})$ that have reached their birth threshold:
\begin{equation}
\begin{aligned}
    V(t) &= \{ v \in \mathcal{V} \mid t \ge \tau_v \}\,, \\
    E(t) &= \{ (v_p, v) \in \mathcal{E} \mid t \ge \tau_v \}.
\end{aligned}
\end{equation}

\subsubsection{Optimization}
We employ a Genetic Algorithm (GA) to optimize the parameter set $\Omega = \{ \mathcal{P}, \Delta\tau_{\mu} \}$, where $\mathcal{P}$ denotes the set of production rules, by minimizing the structural discrepancy at the observed boundary timestamps:
\begin{equation}
\argmin_{\Omega} \sum_{i} \text{Dist}_{\text{path}}(L(t_i; L_{t_0}, \Omega), L_{t_i}),
\end{equation}
where we specifically evaluate the case $i=1$ against the target string $L_{t_1}$ evolved from the source state $L_{t_0}$. $\text{Dist}_{\text{path}}$ is the structural distance defined by the Dice coefficient over the sets of unique positional paths $\mathcal{H}$:
\begin{equation}
\text{Dist}_{\text{path}}(L_A, L_B) = 1 - \frac{2 |\mathcal{H}(L_A) \cap \mathcal{H}(L_B)|}{|\mathcal{H}(L_A)| + |\mathcal{H}(L_B)|}.
\end{equation}
Each path in $\mathcal{H}$ encodes the symbolic identity of a node along with its hierarchical address and sibling order to ensure a topology-aware comparison. To strictly enforce structural identity, we introduce a thresholding gate where the distance is fixed to a small epsilon unless $\mathcal{H}(L(t_i; L_{t_0}, \Omega)) = \mathcal{H}(L_{t_i})$, at which point the discrepancy reaches zero. This yields a stable topological sequence that serves as the prerequisite for subsequent geometric modeling in Riemannian space.



\section{Structure-aware Riemannian Growth Fields}

To ensure geometric consistency and avoid non-physical artifacts, such as discontinuous erratic shrinking, we model morphogenesis as a continuous trajectory within a stratified Riemannian manifold. This geometric flow is governed by the structural scaffold $L(t)$ inferred in the previous section, ensuring that the plant's evolving topology remains tightly coupled with its physical deformation.

\subsection{Plant Riemannian State Space}
We define the state of the plant at time $t$ as $x(t) = (G(t), \Theta(t))$, where the geometric parameters $\Theta(t)$ are structured to ensure developmental inheritance. For each edge $e = (v_p, v)$, the parameters define a relative transformation from the terminal frame of its parent $v_p$:
\begin{align}
& \Theta(t) = \left( \mathbf{T}_{\text{root}}(t), \{ \theta_e(t) \mid e \in E(t) \} \right), \quad \nonumber\\
& \theta_e(t) = (s_e(t), \mathbf{R}_e(t), \mathbf{\kappa}_e(t))\,,
\end{align}
where $\mathbf{T}_{\text{root}}(t) \in SE(3)$ is the global root pose, $s_e(t) = \log \ell_e(t) \in \mathbb{R}$ is the log-length of edge $e$, ensuring $\ell_e > 0$ \cite{priestley1922growth},
$\mathbf{R}_e(t) \in SO(3)$ is the relative rotation with respect to the parent's terminal tangent,
and $\mathbf{\kappa}_e(t) \in \mathbb{R}^k$ represents the B-spline parameters (Hermite Splines) governing local curvature.

The plant evolves within a stratified Riemannian manifold $\mathcal{M}(x(t))$. Rather than a single fixed-dimensional space, $\mathcal{M}(x(t))$ is a collection of strata where each stratum's geometry is induced by the current state $x(t)$. Specifically, the state space at time $t$ is formed by the product of the root pose and the individual edge manifolds:
$\mathcal{M}(x(t)) = SE(3) \times \prod_{e \in E(t)} \left( \mathbb{R} \times SO(3) \times \mathbb{R}^k \right)$.
That is, the ``sequential chain of topological events'' corresponds to the trajectory $x(t)$ transitioning between strata of increasing dimensionality. At each birth threshold $\tau_v$, a new manifold factor $(\mathbb{R} \times SO(3) \times \mathbb{R}^k)$ is appended to the product. Continuity is preserved by initializing these new dimensions at the singular boundary where the edge length is approximated to zero ($s_e \to -\infty$), ensuring the plant geometry unfolds smoothly from the progenitor.

\subsection{Continuous Growth Velocity Field}
\begin{figure}[t]
\centering
\includegraphics[clip, trim=0cm 0cm 0cm 0cm, width=1\linewidth]{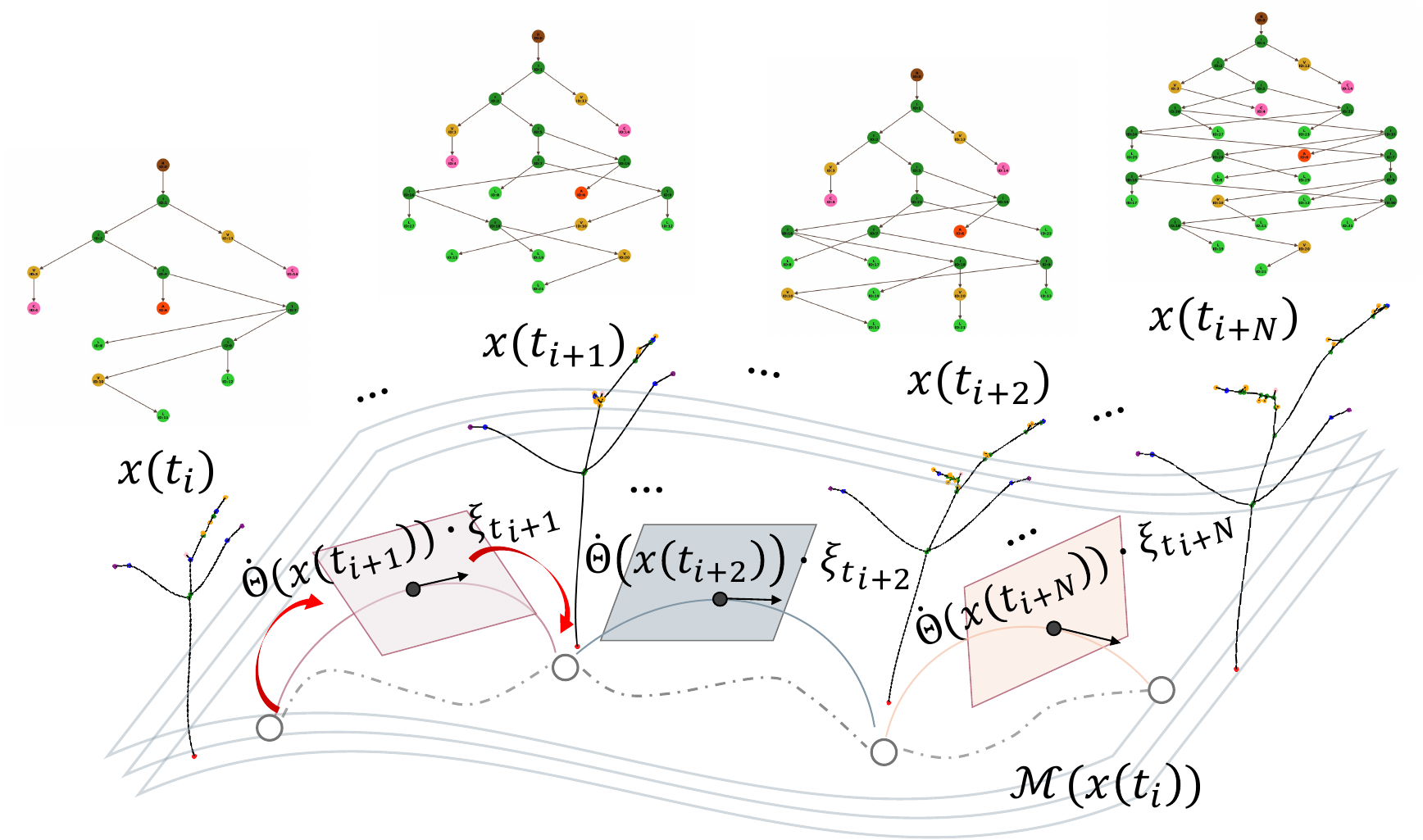}
\caption{Continuous Growth Velocity Field. The developmental trajectory is modeled as a continuous geodesic path across time-varying topological states $x(t)$ on a stratified manifold. The growth intensity, governed by the modulation factor $\xi$, regulates the transition from rapid organ elongation to biological maturation. }
\label{fig:structure_riemannian}
\end{figure}

While the state space $\mathcal{M}(x(t))$ defines the static configurations of the specimen, capturing morphogenesis requires a formal description of how these states evolve over time. To model plant morphogenesis as a continuous biological process, as shown in \figref{fig:structure_riemannian}, we represent growth as a flow generated by a time-dependent velocity field $\dot{\Theta}(x(t), t)$ within the tangent space of $\mathcal{M}(x(t))$. Rather than performing discrete interpolation between topological states, the trajectory $x(t)$ is defined as the integral of this field over the growth interval $[t_0, t_1]$:
\begin{equation}
    x(t_1) = x(t_0) + \int_{t_0}^{t_1} \dot{\Theta}(x(t), t) \, dt \,.
\end{equation}

The velocity vector $\dot{\Theta} = (\dot{\mathbf{T}}_{\text{root}}, \dot{s}_e, \omega_e, \dot{\mathbf{\kappa}}_e)$ captures the expansion and deformation rates of each organ. By parameterizing $\dot{\Theta}$ as a smooth function of time, we ensure that growth remains geometrically consistent even as the underlying graph $G(t)$ undergoes topological transitions. Specifically, at the birth of a new edge $e \in E(t)$, the velocity field handles the activation of its growth potential, ensuring a smooth transition in the physical coordinate space as the organ elongates from its parent.

For each active edge $e$, the growth rates are defined as follows:
\begin{itemize}
    \item Elongation Rate: The term $\dot{s}_e(t) = \frac{d}{dt}\log \ell_e(t)$ represents the \textit{relative growth rate}. In log-space, this linear rate translates to exponential expansion in physical space.
    \item Angular Velocity: The rotational rate $\omega_e(t) \in \mathfrak{so}(3)$ governs the orientation via $\dot{\mathbf{R}}_e(t) = \mathbf{R}_e(t)[\omega_e(t)]_{\times}$, ensuring smooth tropism along manifold geodesics.
    \item Curvature Evolution: The term $\dot{\mathbf{\kappa}}_e(t) \in \mathbb{R}^k$ captures the temporal deformation of the organ's medial axis. Since the B-spline parameters reside in a Euclidean subspace of the stratum, their evolution represents the continuous bending of the organ over time.
\end{itemize}


\subsection{Morphological Smoothness}
Integrating the velocity field yields a family of kinematically valid trajectories; however, additional constraints are required to filter for paths that honor the metabolic and structural limits of real botanical systems.

To ensure the inferred morphogenesis is biologically plausible, we define $E_{\text{smooth}}$ as a penalty on the total variation of the growth velocity field. This term favors trajectories that exhibit steady growth and minimal unnecessary deformation, effectively acting as a ``minimum-jerk'' prior for plant development. We define the smoothness energy as the integral of the squared norm of the manifold acceleration as follows:
\begin{equation}
E_{\text{smooth}} = \int_{t_0}^{t_1} \left\| \frac{D \dot{\Theta}(t)}{dt} \right\|^2_{\mathcal{M}} dt \,,
\end{equation}
where $\frac{D}{dt}$ denotes the covariant derivative. Expanding this into our parameter space, the term penalizes the second-order derivatives of the log-length, orientation, and curvature, ensuring that growth remains steady across the stratified manifold:
\begin{align}
&\left\| \frac{D \dot{\Theta}(t)}{dt} \right\|^2_{\mathcal{M}} = \nonumber \\
&\sum_{e \in G(t)} \left( \alpha_e |\ddot{s}_e(t)|^2 + \beta_e \|\dot{\omega}_e(t)\|^2_{\mathfrak{so}(3)} + \eta_e \|\ddot{\mathbf{\kappa}}_e(t)\|^2 \right) \,,
\end{align}
where $\alpha_e, \beta_e,$ and $\eta_e$ are the semantic-dependent weights representing axial, flexural, and configurational stiffness within the range $[0.1, 1.0]$. In our experiments, we set high penalties ($\alpha_{\text{e}} = \beta_{\text{e}} = \eta_{\text{e}} = 1.0$) for primary structural elements such as internodes to enforce structural rigidity. Conversely, we assign lower weights for terminal organs ($\alpha_e = 0.1$ and $\beta_e = \eta_e = 0.2$) for leaves or petioles to accommodate complex environmental deformations (e.g., phototropic reorientation) without incurring excessive smoothness penalties. 

\paragraph{Sigmoid Growth Modulation.}
We assume that the growth follows a sigmoid characteristic \cite{priestley1922growth}, \D{typically decaying as $\theta_e(t)$ approaches a saturation}. To ensure this property and simultaneously preserve the flexibility of the intrinsic elongation potential, we define the modulation factor $\xi_{\theta_e}$ based on the \D{derivative of the log-sigmoid function}. Here, we introduce the \D{component-specific absolute peak growth timing} $T_{\theta_e} = \tau_v + (1 - \tau_v) t_{m, \theta_e}$, where $t_m \in [0, 1]$ is a global parameter normalized relative to the remaining lifespan from the inception $\tau_v$:
\begin{align}
    \xi(k_{\theta_e}, T_{\theta_e}) & = \frac{d}{dt} \log \left( \frac{1}{1 + e^{-k_{\theta_e}(t-T_{\theta_e})}} \right) \nonumber \\
    & = \frac{k_{\theta_e}}{1 + e^{k_{\theta_e}(t-T_{\theta_e})}}.
\end{align}
The velocity field is then updated as \D{$\dot{\theta}_e(t) \to \dot{\theta}_e(t) \cdot \xi_{\theta_e}$}.
To regularize the trajectory within the energy functional $E_{\text{smooth}}$, the second-order derivative (acceleration) for any growth component $\theta_e(t)$ is given by:
\D{
\begin{equation}
\ddot{\theta}_e(t) \simeq - \dot{\theta}_e(t) \frac{k_{\theta_e}^2 e^{k_{\theta_e}(t - T_{\theta_e})}}{(1 + e^{k_{\theta_e}(t - T_{\theta_e})})^2} \,.
\end{equation}
}
This generalized form ensures that the growth of length, curvature, and torsion are all governed by the same underlying biological saturation logic, while allowing for distinct parameters $\{k, t_m\}$ for each physical property.



\section{Holistic 4D Plant Modeling}

Building upon the structural trajectories, we expand our skeletal manifold into a holistic 4D representation by modeling deformation of stems and leaves across the temporal domain.

\subsection{Deformable Leaf}
We treat each leaf as a deformable template $P_{leaf}(t)$ anchored to a specific node $v$. To maintain consistency with the continuous Riemannian framework, the global trajectory of any point $\mathbf{p}(t)$ on the leaf surface is defined by the composition of the skeletal rigid-body motion and an intrinsic anisotropic expansion flow:
\begin{equation}
\mathbf{p}(t) = \mathbf{x}_{v}(t) + \mathbf{R}_{leaf}(t) \cdot \mathbf{\Lambda}(t) \cdot \mathbf{p}_{local} \,,
\end{equation}
where $\mathbf{x}_{v}(t)$ is the absolute position of the anchor node and $\mathbf{p}_{local}$ is the canonical template point. The leaf-specific orientation $\mathbf{R}_{leaf}(t)$ is governed by the angular velocity $\dot{\omega}_{leaf}(t)$ with sigmoid modulation, coupled with the primary axis connecting the anchor node to the distal leaf tip.

The intrinsic growth of the leaf blade is modeled as an \textit{anisotropic scaling flow} in the local canonical space. We define the deformation tensor $\mathbf{\Lambda}(t) = \text{diag}(e^{s_{long}(t)}, e^{s_{lat}(t)}, e^{s_{lat}(t)})$, where $s_{long}$ and $s_{lat}$ represent the longitudinal (midrib) and latitudinal (lamina) log-scales, respectively. To ensure biological plausibility, these scaling rates are slaved to the corresponding organ-specific sigmoid modulation $\xi(t)$ \D{(Sec.~4.3)}:
\begin{equation}
\begin{bmatrix} \dot{s}_{long}(t) \ \dot{s}_{lat}(t) \end{bmatrix} \rightarrow \begin{bmatrix} \dot{s}_{long} \ \dot{s}_{lat} \end{bmatrix} \cdot \xi(t) \,,
\end{equation}
where $\dot{s}$ represents the constant growth velocity in the log-scale space. 
For leaves emerging post-birth ($\tau_v > t_0$), the scale is initialized to an infinitesimal value $\epsilon$, representing the transition from a dormant bud to an expanding lamina.


\subsection{Deformable Stem}
We model the stem as a continuous assembly of \textit{Generalized Cylinders} parameterized by the underlying skeletal splines \cite{shani1984splines}. To ensure structural continuity at branching junctions, we define the radius $r_v(t)$ at each node $v$ rather than per edge. For an edge $e$ connecting parent $v_p$ to child $v$, the radius at any longitudinal position $u \in [0, 1]$ is determined by the interpolation:
$r_e(u, t) = (1-u)r_{v_p}(t) + u \, r_{v}(t)$.
By inheriting the propagation logic from the skeletal graph, $r_v(t)$ scales from its initial measured value at $t_0$ (or an small value $\epsilon$ for nodes emerging post-birth) toward its terminal thickness at $t_1$.

The stem surface $S_e$ is procedurally generated at time $t$ via:
\begin{align}
S_e(u, \phi, t) = & \gamma_e(u, t) + \nonumber \\ 
& r_e(u, t) \left[ \cos(\phi)\mathbf{N}(u, t) + \sin(\phi)\mathbf{B}(u, t) \right] \,,
\end{align}
where $\gamma_e(u, t)$ is the skeletal Hermite spline, $\phi \in [0, 2\pi)$ is the azimuthal angle, and $\{\mathbf{N}, \mathbf{B}\}$ represents the orthonormal basis of the \textit{Bishop Frame} propagated via parallel transport \cite{bishop1975there}. This procedural mapping ensures that surface points automatically scale with the skeletal growth, eliminating the need for explicit point-level deformation. 


\subsubsection{\D{Optimization}}
By combining these dynamic flow constraints with structural priors, we can formulate a global energy that reconciles the model with observed physical evidence.
The optimization is performed over the control coefficients for all the modulation parameters $\{k, t_m\}$ and is defined as:
\begin{equation}
E(k, t_m) = E_{\text{data}} + \lambda_1 E_{\text{smooth}} \,,
\end{equation}
where $E_{\text{data}} = d(\hat{P}(t),\, P_{t})$ represents the data fidelity term. The function $d(\cdot)$ computes the Chamfer distance between the estimated point cloud $\hat{P}(t)$—integrated forward from the fixed initial state $\Theta(t_0)$—and the corresponding target observation $P_{t}$.



\begin{figure*}[t]
     \centering
     \includegraphics[width=0.90\linewidth]{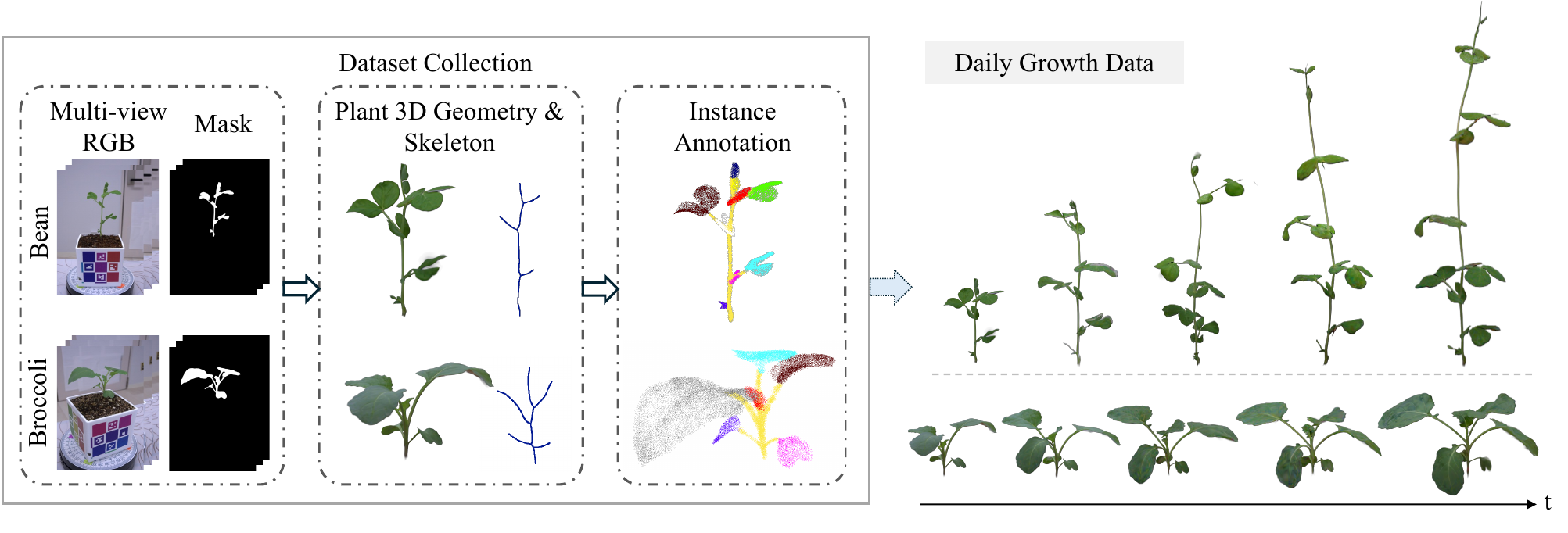}
     \vspace{-10pt}
     \caption{
    \D{Bean and broccoli data collection over 10 days. Each day, for every specimen, we capture about 30 multi-view RGB images, extract masks using SAM 3 \cite{carion2025sam}, reconstruct the 3D geometry through a Gaussian Splatting--based multi-view pipeline, extract the skeleton, and perform instance annotation, yielding the temporal sequence of 3D geometry. The same process generalizes to other species.}
     }
     \label{fig:dataset}
 \end{figure*}

\section{Experiments}
In this section, we provide a comprehensive quantitative and qualitative evaluation of our 4D growth model, assessing its performance in terms of topological stability, morphological fidelity, and temporal correspondence.

\subsection{Real Plant Dataset Acquisition and Processing}
To evaluate our 4D growth model, we utilized both the publicly available Pheno4D tomato dataset \cite{schunck2021pheno4d} and a custom-collected dual-species temporal dataset. \D{As shown in \figref{fig:dataset}}, \D{our dataset consists of daily growth measurements of bean and broccoli specimens over 10 days.}
We acquired about 30 multi-view images per specimen per day using a Sony $\alpha$7 ILCE-7K. We generated sparse 3D reconstructions via 3D Gaussian Splatting (3DGS) \cite{ye2025gsplat}. To ensure geometric density and semantic purity, 3D geometries were projected into 2D space, refined using the Segment Anything Model 3 (SAM 3) \cite{carion2025sam}, and back-projected to 3D. Absolute scale was preserved via ChArUCO marker calibration. \D{Skeletons were extracted using Laplacian-based contraction \cite{meyer2023cherrypicker} and refined via directional search (Sec. 3.1)}. Ground truth semantic labels (stem, leaf, cotyledons) of point clouds were manually annotated 
to provide a baseline for topological recovery.




\subsection{Quantitative Evaluation}
\D{We quantitatively evaluate the modeling accuracy and robustness of our method on both Pheno4D tomato specimens \cite{schunck2021pheno4d} and our multi-species dataset.}
\D{\tabref{tab:quantitative_results} reports our method against skeletal \cite{khanam2024riemannian} and holistic registration \cite{pan2021multi} baselines. Errors are reported as the mean Chamfer Distance (CD) in mm. The results demonstrate the fundamental advantage of our approach in integrating a kinematically valid growth trajectory across time, outperforming baselines that struggle with the large displacements and topological changes inherent in botanical development, especially across distant timepoints.}

\begin{figure}[t]
    \centering
    \begin{minipage}[b]{0.545\linewidth}
        \centering
        \includegraphics[width=\linewidth]{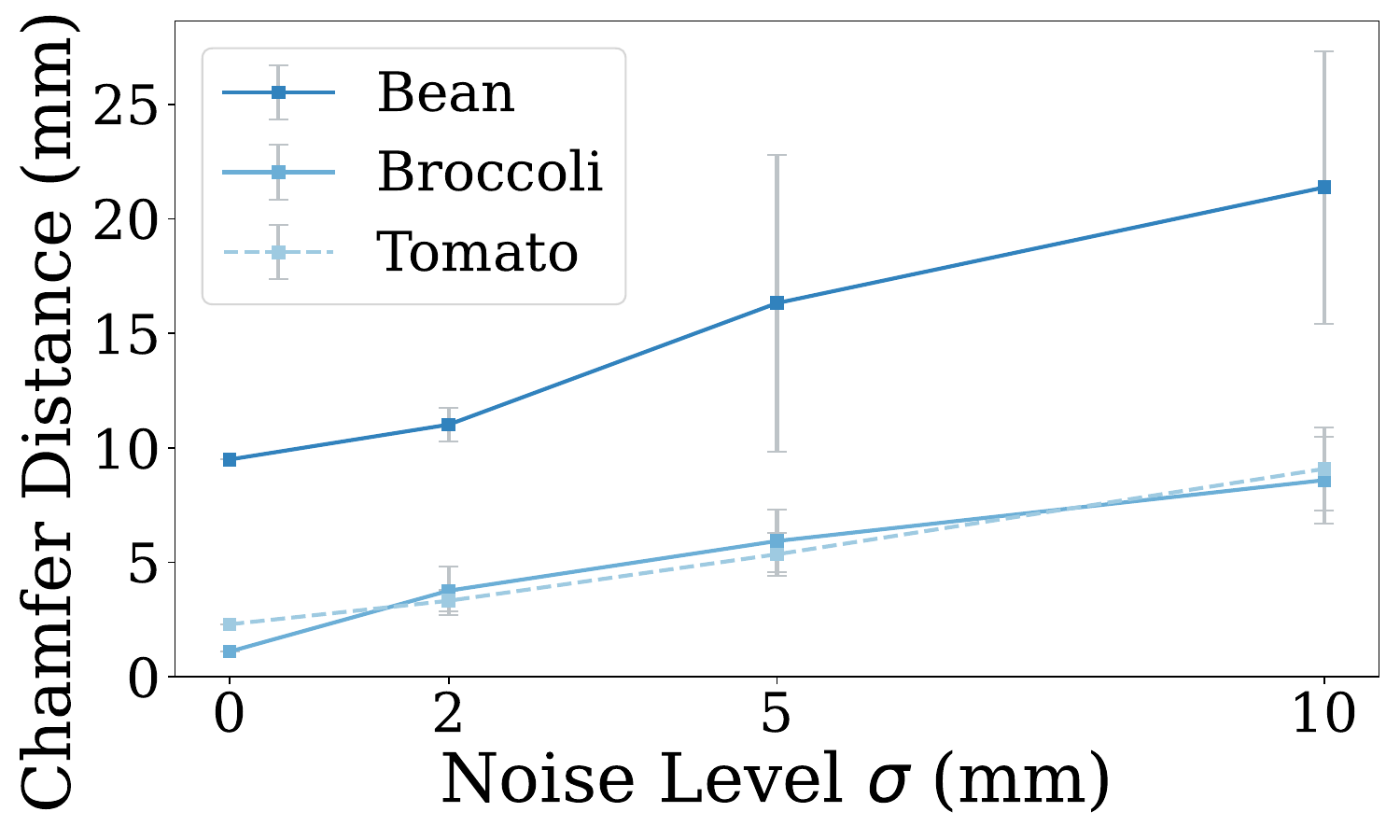}
        \vspace{-1.8em} \\
        {\quad \footnotesize (a)}
    \end{minipage}
    \hfill 
    \begin{minipage}[b]{0.435\linewidth}
        \centering
        \includegraphics[width=\linewidth]{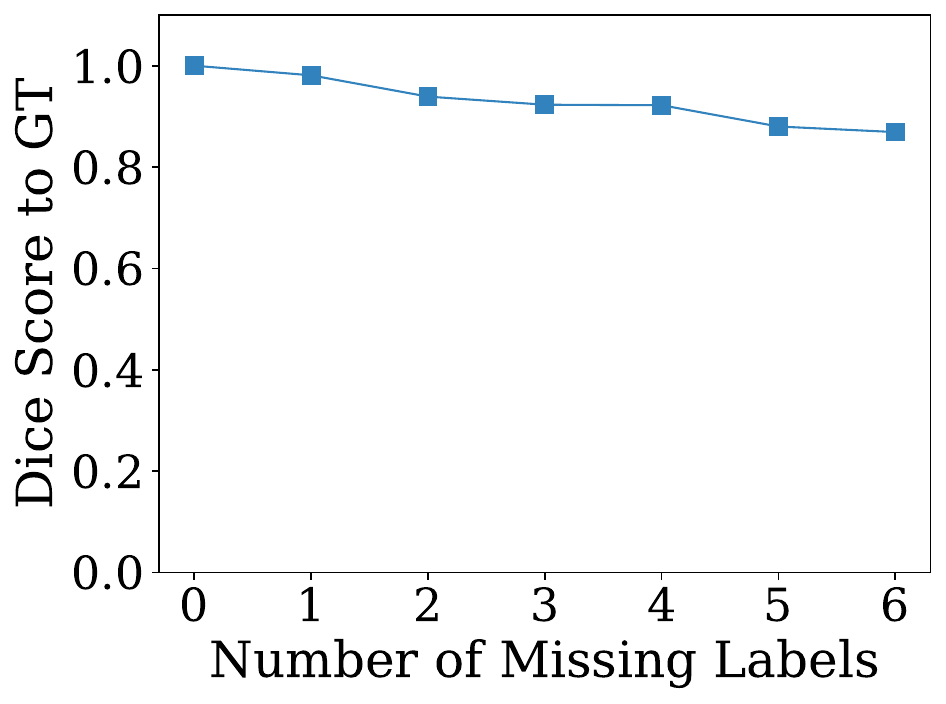}
        \vspace{-1.8em} \\
        {\quad\quad \footnotesize (b)}
    \end{minipage}
    
    \vspace{2pt} 
    
    \begin{minipage}[b]{0.54\linewidth}
        \centering
        \includegraphics[width=\linewidth]{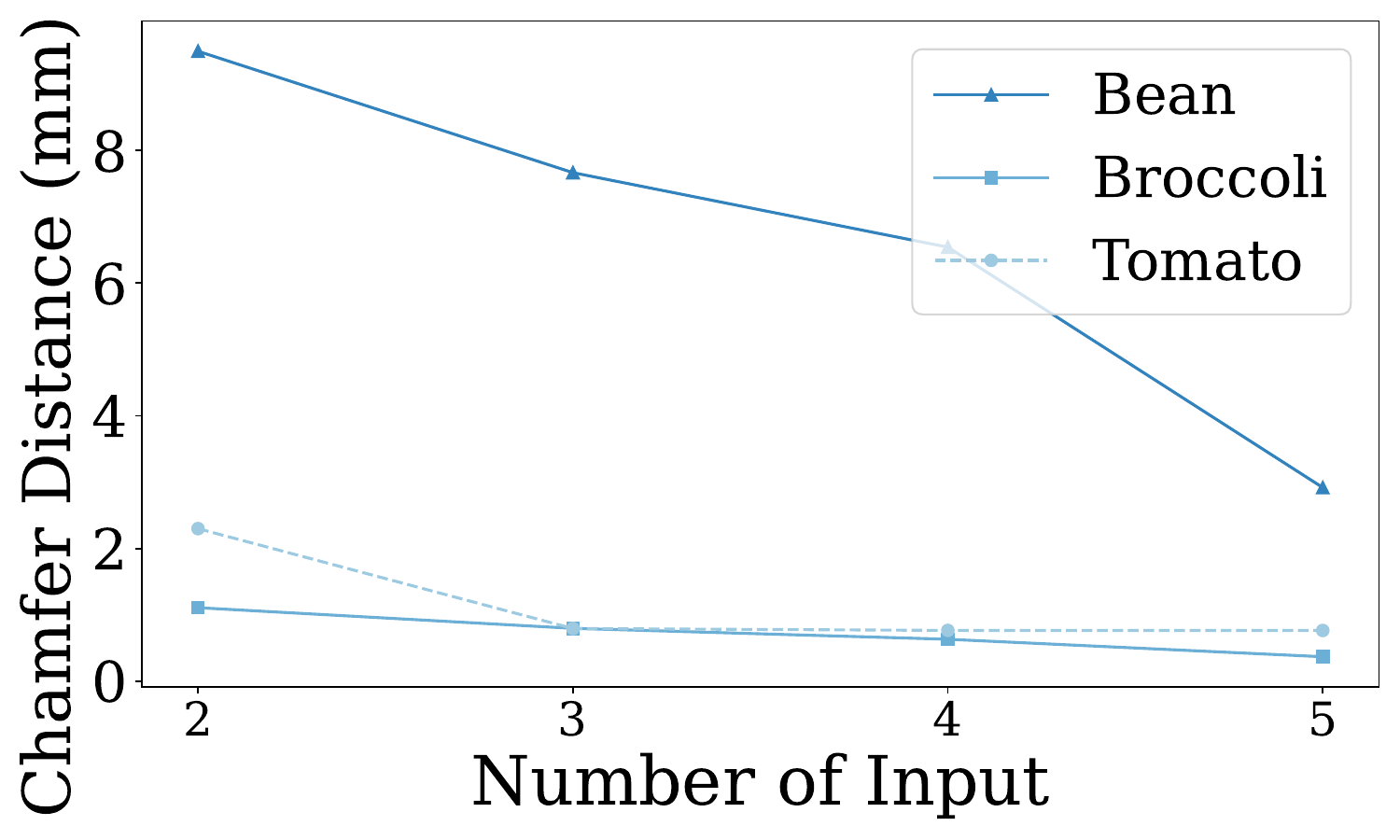}
        \vspace{-1.8em} \\
        {\quad \footnotesize (c)}
    \end{minipage}
    
    \vspace{-0.6em} 
    \caption{
    \D{Robustness analysis. (a) Mean estimation errors at different noise levels. (b) GA robustness under missing semantic labels. (c) Mean estimation errors at different number of inputs. Colors denote different species.}
    }
    \label{fig:noise_plot}
\end{figure}

\D{\figref{fig:noise_plot} presents the robustness analysis. In \figref{fig:noise_plot}(a), we report the mean temporal Chamfer Distance under Gaussian noise ($\sigma \in [0, 10]$~mm) injected into the input point-cloud pairs for broccoli, bean, and tomato; error increases linearly with noise magnitude. \figref{fig:noise_plot}(b) evaluates the GA-recovered rules under input corruption: on a tomato with 11 leaves, we randomly drop leaf labels at both timepoints and report the mean structural error via the Dice coefficient (higher is better) against ground truth. Performance degrades smoothly as corruption increases. Moreover, in \figref{fig:noise_plot}(c), we report the estimation error for different numbers of input observations. The results show that modeling accuracy improves as the number of inputs increases.}
\D{\tabref{tab:ablation_results} investigates the impact of the sigmoid modulation factor $\xi$ by comparing our full model against a baseline that assumes a constant growth rate. Our full model reduces errors across all species, validating the effectiveness of the modulation.}



\begin{table}[t]
\vspace{-3pt}
\centering
\caption{\D{Quantitative comparison of 4D reconstruction accuracy. We report Skeletal Chamfer Distance (Sk.) and holistic surface CD (Hl.) in mm. Note that \cite{khanam2024riemannian} reports only skeletal results, while \cite{pan2021multi} reports only holistic results. Our method achieves higher accuracy.}}
\label{tab:quantitative_results}
\resizebox{\linewidth}{!}{\begin{tabular}{@{}l ccc@{}}
\toprule
 \textbf{Method} & \multicolumn{1}{c}{\textbf{Tomato} $\downarrow$} & \multicolumn{1}{c}{\textbf{Bean} $\downarrow$} & \multicolumn{1}{c}{\textbf{Broccoli} $\downarrow$}  \\
\midrule
Pan \etal \cite{pan2021multi} Hl.
& 12.68 & 16.11  & 6.32 \\
Khanam \etal \cite{khanam2024riemannian} Sk.
& 21.12 & 20.23 & 2.09 \\
\textbf{Ours} Sk.
&  \D{\textbf{2.09}} &  \D{\textbf{8.83}} &  \D{\textbf{1.01}}  \\
\textbf{Ours} Hl.
&  \D{\textbf{5.45}} &  \D{\textbf{7.86}} &  \D{\textbf{1.92}}  \\
\bottomrule
\end{tabular}
}
\vspace{-5pt}
\end{table}

\begin{table}[t]
\centering
\caption{\D{Ablation study on geometric priors}. We report mean Skeletal (Sk.) and Holistic (Hl.) Chamfer Distances (mm) w/ and w/o Sigmoid Modulation $\xi$.}
\label{tab:ablation_results}
\setlength{\tabcolsep}{4pt} 
\begin{tabular}{@{}l ccc ccc@{}}
\toprule
& \multicolumn{2}{c}{\textbf{Tomato}} & \multicolumn{2}{c}{\textbf{Bean}} & \multicolumn{2}{c}{\textbf{Broccoli}} 
\\
\cmidrule(lr){2-3} \cmidrule(lr){4-5} \cmidrule(lr){6-7}
\textbf{Method} & Sk. $\downarrow$ & Hl. $\downarrow$ & 
Sk. $\downarrow$ & Hl. $\downarrow$  & 
Sk. $\downarrow$ & Hl. $\downarrow$ \\ 
\midrule
w/o $\xi$
& 2.30 & 5.85
& \D{9.49} & 8.44
& \D{1.11}  & 2.01  
 \\
\textbf{Ours} 
& \D{\textbf{2.09}} & \D{\textbf{5.45}}
& \D{\textbf{8.83}} & \textbf{7.86}  
& \D{\textbf{1.01}} & \textbf{1.92} 
 \\
\bottomrule
\end{tabular}
\vspace{-5pt}
\end{table}


\subsection{Qualitative Evaluation}
\begin{figure}[t]
     \centering
     \includegraphics[width=\linewidth]{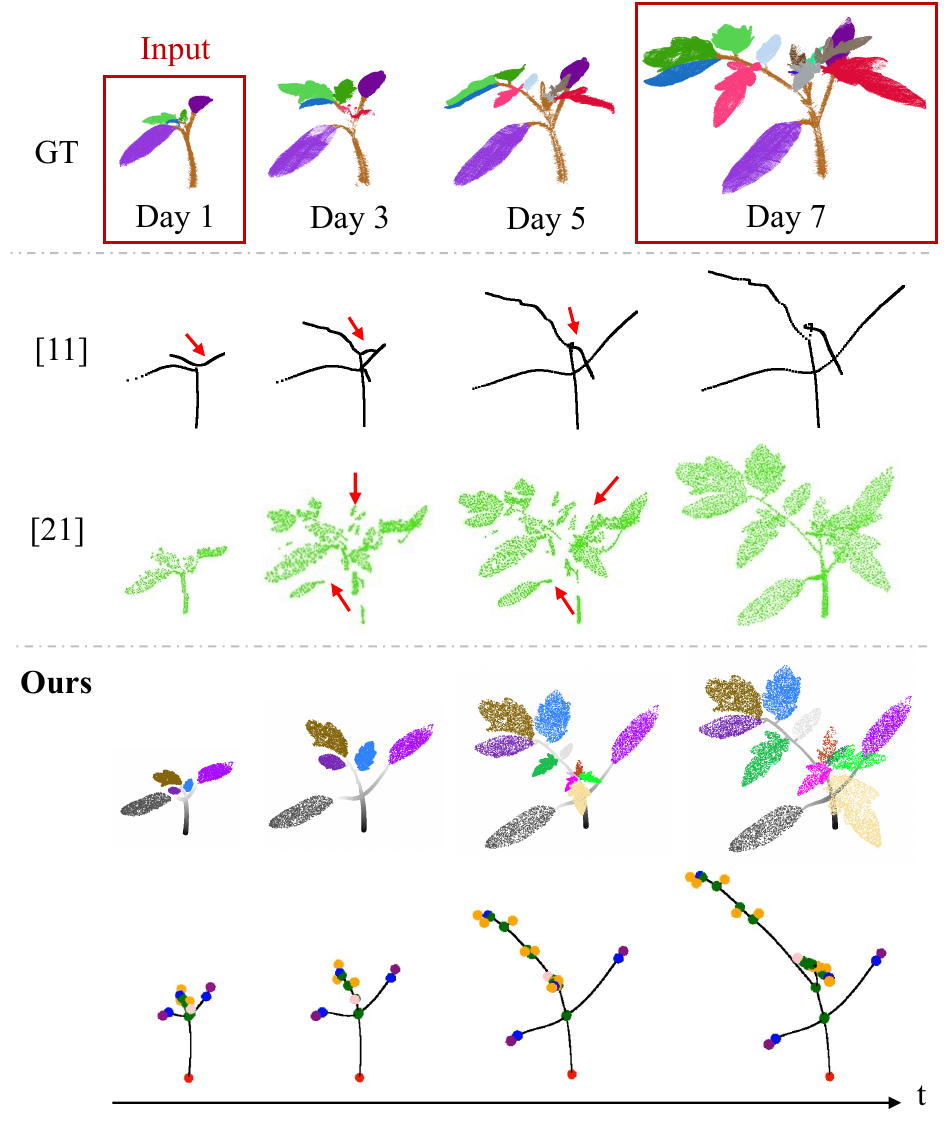}
     \vspace{-14pt}
     \caption{\D{Qualitative 4D Reconstruction of Pheno4D Tomato. We show recovered growth trajectories of a tomato specimen over time. The input pair is highlighted with red boxes. From top to bottom, the rows display: GT temporal 3D point clouds; skeletal reconstruction from Khanam \etal \cite{khanam2024riemannian}; spatiotemporal registration results from Pan \etal \cite{pan2021multi}; our generated continuous holistic geometry; and our recovered topological skeletons with invariant node identities (indicated by consistent color-coding). Our approach better recovers temporally continuous geometry.}}
     \label{fig:qualitative_tomato}
     \vspace{-5pt}
 \end{figure}

\begin{figure}[t]
    \centering
    \includegraphics[clip, trim=0cm 0cm 0cm 0cm, width=1\linewidth]{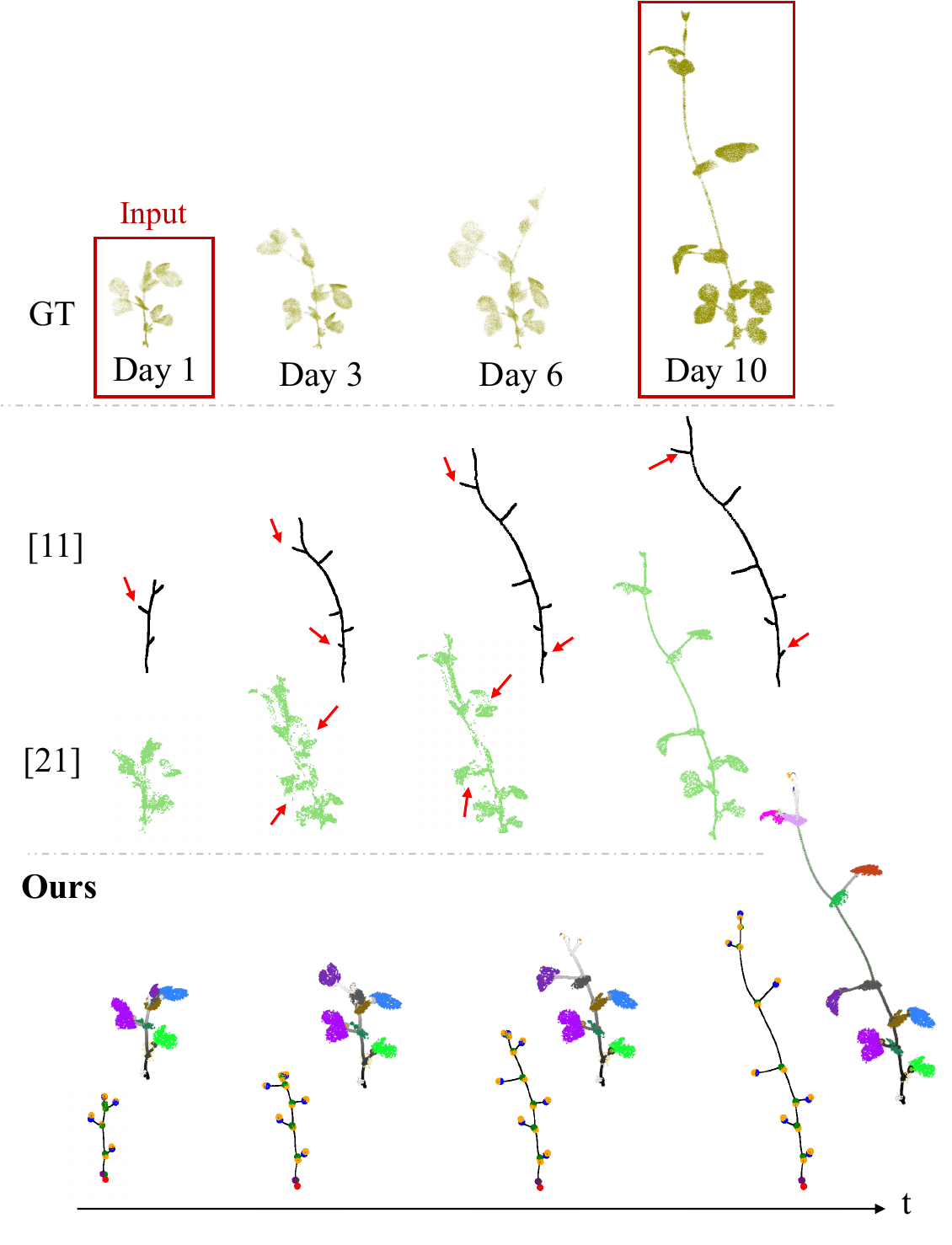}
    \vspace{-14pt}
    \caption{\D{Qualitative 4D Reconstruction of Bean. We visualize the recovered growth trajectories of bean specimens over time. The input pair is highlighted with red boxes. Note that our Riemann flow preserves better branch topology and identity.}
    }
    \label{fig:qualitative_bean}
\end{figure}

\begin{figure}[t]
     \centering
     \includegraphics[width=\linewidth]{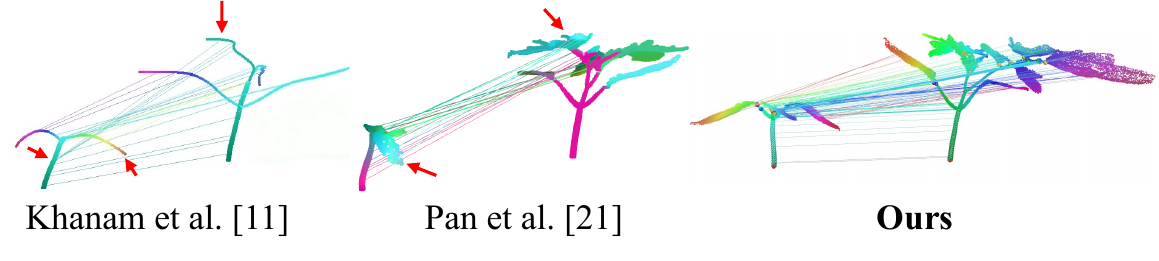}
     \vspace{-15pt}
     \caption{
    \D{Qualitative comparison of dense correspondences recovered from two sparse timepoints (matching colors denote identical points). Our method yields markedly more consistent correspondences, whereas baselines produce non-corresponding matches.}
     }
     \label{fig:qualitative_corresp}
     \vspace{-5pt}
 \end{figure}

\D{As shown in \figref{fig:qualitative_tomato} and \figref{fig:qualitative_bean}, our model accurately captures the non-linear morphological progression of tomato and bean specimens. The continuous Riemannian flow preserves surface manifold integrity across all observed time instances. In contrast, baseline methods \cite{pan2021multi, khanam2024riemannian} either exhibit tracking drift, where branches slide laterally or translate upward, or produce fragmented intermediate geometry when interpolating between distant snapshots (indicated by red arrows).} 
\D{This structural stability is particularly vital during organ inception. By initializing growth at the singular boundary of the stratified manifold, our model ensures that new leaves and internodes unfold smoothly from their progenitors, eliminating the discrete popping or volume loss. \figref{fig:qualitative_corresp} compares dense correspondences recovered from two sparse timepoint observations. While our method yields consistent dense correspondences, the baselines produce spurious matches that erroneously link non-corresponding points, validating our approach's ability to accurately track growth trajectories across development.}

\section{Conclusion}
In this paper, we presented a novel framework for reconstructing continuous 4D plant morphogenesis from sparse temporal observations. By formulating plant development as geodesic evolution on a structure-aware Riemannian growth field, our method jointly models topology evolution and geometric deformation within a single continuous representation, enabling stable dense correspondences and precise organ-level growth tracking. Experimental result demonstrate the effectiveness of our approach in capturing complex growth dynamics.
Yet, challenges remain — severe self-occlusions and reconstruction noise can affect reconstruction stability. Moreover, fine-scale phenomena such as leaf unflurring or subtle surface unfolding are not explicitly modeled. Addressing these limitations through tighter integration of biomechanical and appearance-aware dynamics remains an important direction for future work. Despite these challenges, we believe our formulation establishes a principled foundation for 4D plant modeling and future predictive digital plant twins.

\vspace{-5pt}
\paragraph*{\D{Acknowledgment}} \D{This work was in part supported by the Organization for the Promotion of Gender Equality at Nara Women's University.}

{
    \small
    \bibliographystyle{ieeenat_fullname}
    \bibliography{main}
}

\end{document}